\documentclass[sigconf,nonacm]{acmart}
\usepackage{hyperref}
\usepackage{algorithm}
\usepackage{algpseudocode}
\usepackage{pgfplots}
\pgfplotsset{compat=1.17}

\AtBeginDocument{%
  \providecommand\BibTeX{{%
    \normalfont B\kern-0.5em{\scshape i\kern-0.25em b}\kern-0.8em\TeX}}}

\setcopyright{none}
\begin{document}

\title{Evolutionary Policy Optimization}

%\author{Zelal Su (Lain) Mustafaoglu}
%\email{zsm@utexas.edu}
%\affiliation{%
  %\institution{University of Texas at Austin}
  %\city{Austin}
  %\state{TX}
  %\country{USA}
%}

\author{Zelal Su (Lain) Mustafaoglu}
\authornote{Corresponding author: Lain Mustafaoglu, Email:zsm@utexas.edu}
\author{Keshav Pingali}
\author{Risto Miikkulainen}
\affiliation{%
  \institution{University of Texas at Austin}
  \city{Austin}
  \state{TX}
  \country{USA}
}


\begin{abstract}
    A key challenge in reinforcement learning (RL) is managing the exploration-exploitation trade-off without sacrificing sample efficiency. Policy gradient (PG) methods excel in exploitation through fine-grained, gradient-based optimization but often struggle with exploration due to their focus on local search. In contrast, evolutionary computation (EC) methods excel in global exploration, but lack mechanisms for exploitation. To address these limitations, this paper proposes Evolutionary Policy Optimization (EPO), a hybrid algorithm that integrates neuroevolution with policy gradient methods for policy optimization. EPO leverages the exploration capabilities of EC and the exploitation strengths of PG, offering an efficient solution to the exploration-exploitation dilemma in RL. EPO is evaluated on the Atari Pong and Breakout benchmarks. Experimental results show that EPO improves both policy quality and sample efficiency compared to standard PG and EC methods, making it effective for tasks that require both exploration and local optimization.
\end{abstract}

\begin{comment}
    
\begin{CCSXML}
<ccs2012>
 <concept>
  <concept_id>10010520.10010553.10010562</concept_id>
  <concept_desc>Computer systems organization~Embedded systems</concept_desc>
  <concept_significance>500</concept_significance>
 </concept>
 <concept>
  <concept_id>10010520.10010575.10010755</concept_id>
  <concept_desc>Computer systems organization~Redundancy</concept_desc>
  <concept_significance>300</concept_significance>
 </concept>
 <concept>
  <concept_id>10010520.10010553.10010554</concept_id>
  <concept_desc>Computer systems organization~Robotics</concept_desc>
  <concept_significance>100</concept_significance>
 </concept>
 <concept>
  <concept_id>10003033.10003083.10003095</concept_id>
  <concept_desc>Networks~Network reliability</concept_desc>
  <concept_significance>100</concept_significance>
 </concept>
</ccs2012>
\end{CCSXML}

\ccsdesc[500]{Computer systems organization~Embedded systems}
\ccsdesc[300]{Computer systems organization~Redundancy}
\ccsdesc{Computer systems organization~Robotics}
\ccsdesc[100]{Networks~Network reliability}
\end{comment}


\maketitle

\section{Introduction}
A fundamental problem in deep reinforcement learning (RL) is the exploration vs. exploitation trade-off. Agents must balance the exploitation of good actions with exploration to discover better actions. This trade-off becomes particularly challenging in environments with sparse rewards, deceptive local optima, or large state-action spaces. A crucial concern for all learning methods is their {\em sample complexity} - that is, the number of interactions with the environment they require to learn a near-optimal policy with a certain level of accuracy. This is important because in practical problems, each interaction requires the expenditure of time and resources, and a method that is sample inefficient may have limited practical utility. 

Gradient-based approaches excel at exploitation by incrementally improving policies through gradient-based updates. Policy gradient (PG) methods optimize the parameters of the policy network with gradient ascent. Advanced PG methods, such as proximal policy optimization (PPO) \cite{SchulmanPPO}, aim to increase sample efficiency and stability of vanilla PG methods by limiting the size of gradient updates and reusing samples. However, these methods  rely on local, conservative updates, and this often limits their ability to explore beyond the current policy distribution and escape from local minima. 

In contrast, gradient-free optimization methods, such as evolutionary computation (EC), offer an alternative approach by focusing on global exploration of the parameter space. This makes gradient-free methods particularly suitable for tasks with sparse rewards. Gradient-free approaches like the cross-entropy method (CEM) \cite{RubinsteinCEM} frequently outperform gradient-based approaches.

This raises a natural question: can we combine the strengths of gradient-based and gradient-free methods to achieve effective exploitation and exploration?

{\em This paper provides a full description of Evolutionary Policy Optimization (EPO) \cite{mustafaoglu2025epo}, a method first introduced in our work submitted to GECCO 2025 to address this question. Evolutionary Policy Optimization (EPO) integrates PG methods with an evolutionary approach to boost exploration.}

EPO improves over standalone PG or EC methods by leveraging the fine-grained optimization capabilities of gradient-based approaches while introducing evolutionary operations for exploration. PPO is used to pre-train the initial population of agents, and also to fine-tune individuals created during the evolution process. 

EPO is evaluated on two Atari benchmarks, Pong and Breakout, spanning a range of reward densities and task complexities. On Breakout, it improves sample efficiency by 26.8\% over PPO and 57.3\% over pure evolution, while in Pong, it discovers significantly better policies. These results highlight the benefits of combining neuroevolution with policy gradients, making EPO a strong approach for RL tasks requiring both structured exploration and local optimization.

\section{Background}

Reinforcement learning (RL) formalizes the problem of sequential decision-making, where an agent learns to maximize cumulative rewards through interactions with an environment.  These interactions are typically modeled as a Markov Decision Process (MDP), defined by a state space \(S\), an action space \(A\), transition dynamics \(P(s'|s, a)\) (which models the influence of the environment), a reward function \(R(s, a)\), and a discount factor \(\gamma \in [0, 1]\). At each timestep \(t\), the agent observes a state \(s_t \in S\), selects an action \(a_t \in A\) according to a policy \(\pi(a|s)\), transitions to a new state \(s_{t+1} \sim P(s'|s_t, a_t)\), and receives a scalar reward \(r_t = R(s_t, a_t)\). 

The goal in RL is to find an optimal policy \(\pi^*\) that maximizes the total expected discounted return:

\[
G_t = \mathbb{E}_{\pi} \left[ \sum_{k=0}^\infty \gamma^k r_{t+k} \right],
\]
where the expectation is taken over the trajectory distribution induced by \(\pi\) and \(P\). The optimal policy \(\pi^*\) maximizes \(G_t\) for all states \(s_t \in S\).

Deep RL extends traditional RL by parameterizing the policy using neural networks and directly optimizing policies via gradient ascent. Parameterizing policies allows agents to operate in environments with high-dimensional or continuous state spaces, where handcrafted features or lookup tables are impractical. The policy distribution \( \pi_{\theta}(a | s) \), parameterized by \(\theta \in \mathbb{R}^d\), maps states to a distribution over actions and is optimized iteratively.

A key challenge in RL is balancing exploration and exploitation: the agent must explore the environment to discover strategies that yield higher rewards, while also exploiting its current knowledge to maximize performance. This challenge is further compounded by the need to optimize policies efficiently, as interactions with the environment are often costly. 

Gradient-based RL approaches, such as policy gradient methods, leverage gradients of expected returns to iteratively improve parameterized policies. In contrast, gradient-free alternatives such as neuroevolution employ evolutionary principles to optimize policies through population-based exploration. These approaches differ significantly in their treatment of exploration and exploitation, and in their sample efficiency. These approaches are described in the following subsections.

\subsection{Policy Gradient Methods}
Policy gradient (PG) methods are a class of reinforcement learning algorithms that directly optimize the policy \( \pi_{\theta}(a | s) \), which is parameterized by \(\theta\) (e.g., the weights of a neural network) \cite{Sutton1999}. Unlike value-based methods, which estimate the value of states or state-action pairs, PG methods aim to optimize a parameterized policy by maximizing the expected cumulative reward.

The objective for PG methods is given by:

\[
J(\theta) = \mathbb{E}_{\tau \sim \pi_{\theta}} \left[ R(\tau) \right] = \sum_{\tau} P(\tau; \theta) R(\tau),
\]
where:
\begin{itemize}
    \item \( \theta \): the parameters of the policy distribution \( \pi_{\theta} \), such as the weights of a neural network.
    \item \( \tau = (s_0, a_0, s_1, \dots, s_T) \): A trajectory or sequence of states \(s_t\) and actions \(a_t\) sampled from the policy \( \pi_{\theta} \), where T is the length of the trajectory. 
    \item \( P(\tau; \theta) \): The probability of a trajectory \(\tau\) under the policy distribution \( \pi_{\theta} \),
    \item \( R(\tau) = \sum_{t=0}^T \gamma^t r_t \): The discounted cumulative reward of the trajectory \(\tau\), where \(r_t\) is the reward received at time \(t\), and \(\gamma \in [0, 1]\) is the discount factor.
\end{itemize}

The policy gradient theorem is used to compute the gradient of the objective \( J(\theta) \), enabling updates to the policy parameters \(\theta\) via gradient ascent. The gradient is given by:

\[
\nabla_{\theta} J(\theta) = \mathbb{E}_{\tau \sim \pi_{\theta}} \left[ \sum_{t=0}^T \nabla_{\theta} \log \pi_{\theta}(a_t | s_t) R_t \right]
\]

REINFORCE, which is one of the earliest policy gradient algorithms, uses Monte Carlo sampling for gradient estimation to optimize policy parameters \cite{WilliamsReinforce}. Subsequent policy gradient methods, such as A2C and A3C, introduced parallelized training to improve sample efficiency and stabilize learning \cite{MnihA2C&A3C}. Soft Actor-Critic (SAC) integrated entropy regularization to encourage exploration while maintaining sample efficiency \cite{HaarnojaSAC}. By maximizing a trade-off between expected return and policy entropy, SAC achieves robust performance in continuous control tasks.

Another class of advanced policy gradient methods is trust region methods, which ensure that policy updates do not deviate too much from the current policy. This promotes stability and improves sample efficiency. ACER constrains the updated policy to not deviate from the average policy and uses truncation strategies to address the stability issues inherent in off-policy actor critics \cite{WangACER}. 
Trust Region Policy Optimization (TRPO) \cite{SchulmanTRPO} enforces a KL divergence constraint on gradient updates to keep the updated policy close to the current policy. The TRPO objective is:

\begin{displaymath}
\max_{\theta} \mathbb{E}_{s \sim \pi_{\theta_\text{old}}} \left[ \frac{\pi_{\theta}(a|s)}{\pi_{\theta_\text{old}}(a|s)} A^{\pi_{\theta_\text{old}}}(s, a) \right] \quad
\end{displaymath}
 
\begin{displaymath}
\text{subject to} \quad \mathbb{E}_{s \sim \pi_{\theta_\text{old}}} \left[ D_{\text{KL}}(\pi_{\theta_\text{old}} \| \pi_{\theta}) \right] \leq \delta,
\end{displaymath}
where \(\frac{\pi_{\theta}(a_t|s_t)}{\pi_{\theta_\text{old}}(a_t|s_t)}\) is the probability ratio between the new and old policies, \(A^{\pi_{\theta_\text{old}}}(s, a)\) is the advantage function, and \(\delta\) is the KL divergence constraint. 

TRPO is computationally intensive due to its reliance on constrained optimization, which requires second-order gradient evaluations and makes it impractical for use in high-dimensional or complex environments. This led to the development of PPO, which uses a clipped surrogate objective to avoid these problems. The PPO objective is:
\begin{displaymath}
L^{\text{PPO}}(\theta) = \mathbb{E}_{t} \left[ \min \left( r_t(\theta) A_t, \text{clip}(r_t(\theta), 1 - \epsilon, 1 + \epsilon) A_t \right) \right],
\end{displaymath}
where \(r_t(\theta) = \frac{\pi_{\theta}(a_t|s_t)}{\pi_{\theta_\text{old}}(a_t|s_t)}\) is the probability ratio between the new and old policies,  restricted to the range \([1 - \epsilon, 1 + \epsilon]\), where \(\epsilon\) is a hyperparameter, and \(A_t\) is the advantage function.

In this paper, we use PPO as the policy gradient method in our EPO implementation because of its status as the state-of-the-art method in RL. 

PPO’s clipping mechanism prioritizes safety by preventing large policy updates. This ensures stable training but limits exploration, as policies tend to stay close to their current distribution. While entropy incentives encourage exploration, they often fail to drive substantial policy shifts needed to discover novel solutions. This is particularly problematic in environments with sparse rewards or deceptive local optima. This limitation demonstrates the need for a method that is good at exploration, such as an evolutionary approach. 

\subsection{Neuroevolution}
Evolutionary computation (EC) methods offer a gradient-free alternative to gradient-based policy gradient methods. 

Evolutionary computation methods can be used to evolve neural networks, and this is referred to as neuroevolution. The weights, weights and topologies \cite{RistoNEAT}, or even the components and hyperparameters of neural networks can be evolved \cite{RistoDeepNEAT}. Evolutionary methods enable the exploration of novel neural networks instead of incremental changes to an existing network.

Neuroevolution has been effective in evolving policy networks for problems where gradient-based approaches struggle due to poor exploration. However, evolution alone is inefficient at fine-tuning policy parameters. 

{\em A natural solution, explored in this paper, is to combine the strengths of both approaches, leveraging policy gradients for exploitation while using neuroevolution to drive exploration.}

\section{Previous Work}

This section describes previous work in evolutionary strategies for policy optimization.

Salimans et al. (2017) proposed Evolution Strategies (ES) as a gradient-free, scalable alternative to reinforcement learning (RL) for policy optimization. ES methods, specifically Natural Evolution Strategies (NES) \cite{Salimans2017EvolStrat}, evolve policies by maintaining a distribution of parameter vectors and using Gaussian noise to add perturbations to policy parameters for exploration. The population is treated as a distribution characterized by a parameter vector, and the gradient of the expected reward is approximated using sampled perturbations. Results on high-dimensional tasks such as Atari and MuJoCo demonstrate that ES is competitive with standard RL algorithms in terms of performance and scalability.

Such et al. (2018) built on Salimans et al. by introducing distributed genetic algorithms (GAs) with novelty search for deep RL \cite{Such2018}. Neural network parameters were encoded using deterministic seeds combined with mutation histories, significantly reducing memory usage and communication overhead in distributed settings. Their framework incorporates novelty search to reward agents for exploring previously unvisited states. Originally introduced in Lehman et al., novelty search focuses on encouraging the exploration of new behaviors rather than optimizing a single objective, laying the foundation for quality diversity techniques \cite{LehmanNoveltySearch}. Experiments on benchmarks such as Atari, maze navigation, and humanoid locomotion show that the GA-based approach performs competitively with gradient-based RL methods, particularly in exploration-heavy tasks.

Conti et al. (2018) built upon the evolution strategies work of Such et al. by integrating Novelty Search (NS) and Quality Diversity (QD) methods to improve exploration in sparse-reward environments \cite{Conti2018}. Novelty Search rewards agents for discovering new behaviors by encouraging exploration of unvisited states, while Quality Diversity combines novelty and task performance to balance exploration and exploitation.
 
First introduced by Francon et al. in 2020, Evolutionary Surrogate-assisted Prescription (ESP) is a framework for optimizing decision strategies by combining surrogate modeling with evolutionary search \cite{FranconESP}. ESP comprises two components: the Predictor, a machine learning model trained on historical data that acts as a surrogate for policy evaluation, and the Prescriptor, a policy network optimized through evolution that maps contexts to optimal actions to maximize long-term rewards as predicted by the surrogate Predictor, which approximates reward functions akin to Q-values. The ESP framework was applied to a variety of RL tasks, ranging from function approximation to Cart Pole, demonstrating its effectiveness in optimizing policies without direct gradient computation.

\textbf{[NEED TO ADD CITATIONS]}

All of this work treats evolution as an {\em alternative} to reinforcement learning. In contrast, the proposed method, Evolutionary Policy Optimization (EPO), {\em integrates} both approaches, leveraging neuroevolution for exploration and policy gradients for efficient refinement. This combination enables more robust learning in environments where previous methods run into difficulties because of limited exploration.

\section{The EPO Algorithm}
\label{sec:epo}

\begin{table*}
    \caption{Hyperparameters for Evolutionary Policy Optimization}
    \label{tab:hyperparameters}
    \centering
    \begin{tabular}{lccc}
        \toprule
        \textbf{Hyperparameter} & \textbf{Description} & \textbf{Search Range} & \textbf{Optimal Value} \\ 
        \midrule
        Mutation Probability & The likelihood of mutating each offspring. & [0.1, 0.5] & 0.3 \\ \hline
        Elite Count & Number of top-performing agents retained per generation. & [2, 6] & 3 \\ \hline
        Population Count & Total agents in each generation. & [6, 16] & 8 \\ 
        \bottomrule
    \end{tabular}
\end{table*}

\begin{comment}
\begin{figure*}
    \centering
\includegraphics[width=0.75\linewidth]{epo-method.png}
    \caption{EPO method --- to be updated with a refined version}
    \label{fig:epo-method}
\end{figure*}
\end{comment}
 
Evolutionary Policy Optimization (EPO) is a hybrid approach that alternates between exploration with neuroevolution and exploitation with PPO. In particular, EPO leverages PPO for policy initialization and for locally optimizing policies that evolution selects through population-based search.

At the start of the algorithm, a policy network is initialized randomly and then pre-trained using PPO for a small number of timesteps. The initial population of agents is then created by cloning this {\em base model} some number of times (hyperparameter $E$). Evolutionary operations are applied as follows. In each generation, the fitness scores of the agents in the population, derived from the reward, are evaluated. To retain the best-performing policies, elite agents are carried over to the new population and act as parents for the next generation. After crossover is applied to the weights of two randomly selected elite agents, mutation is applied to a subset of the offspring, and the other offspring are optimized locally with PPO. The resulting offspring are added to the new population, and this process is repeated until a satisfactory solution is reached. Subsequent sections describe the local optimization and evolution processes in more detail. The pseudocode for EPO is provided in Algorithm~\ref{alg:evo-ppo}. 

%EPO operates through two main steps: 
%\begin{enumerate} 
    %\item Pre-training an initial policy using PPO for informed initialization. 
    %\item Applying evolutionary operations such as elitism, crossover, and mutation to encourage exploration, while exploiting high-performing policies by locally optimizing a subset of offspring with PPO.
%\end{enumerate}

\begin{algorithm}[h]
    \caption{The EPO Algorithm}
    \label{alg:evo-ppo}
    \begin{algorithmic}
        \State Initialize environment $env$, base PPO model $base\_model$, and hyperparameters $m$ ( mutation probability), $P$ (population size), $E$ (elite count), and $T$ (total runtime). 
        \State Pre-train $base\_model$ neural network for $30,000$ timesteps
        \State Initialize population with 2 clones of $base\_model$
        
        \While{Elapsed time $< T$}
            \State Evaluate all agents in the population and compute fitness scores
            \State Retain $E$ top-performing agents (elites)
            
            % \If{Global best fitness improves}
            %     \State Update global best agent
            % \EndIf
            
            \While{$P - E > 0$}
                \State Select two parents $p_1$, $p_2$ randomly from elites
                \State Compute fitness-based crossover scaling factor $\alpha$
                \State Perform crossover for each weight $w$ in child network
                \If{random mutation occurs ($\text{rand()} < m$)}
                    \State Compute mutation scaling factor
                    \State Inject Gaussian noise to weights
                \Else
                    \State Fine-tune offspring for $500$ timesteps
                \EndIf
                \State Add offspring to the next generation
            \EndWhile
            \State Form next generation: $population \gets elites + offspring$
        \EndWhile        
    \end{algorithmic}
\end{algorithm}

\subsection{Evolutionary Operations}

EPO applies three evolutionary operations: elitism, crossover, and mutation. 

\subsubsection{Elitism}

Elitism is a key component of the EPO algorithm as it enables the exploitation promising agents that are discovered through PPO. Elitism ensures that the highest-performing agents are preserved across generations to maintain a baseline of strong policies throughout evolution. The fitness of the agents is determined by the average reward they obtain when evaluated over five episodes. After evaluation, the individuals with the highest fitness values are selected as elite agents. These elites serve two purposes: they are directly carried over to the next generation to exploit promising policies, and they are cloned to serve as parents for generating offspring through crossover and mutation. 

\subsubsection{Crossover}

Crossover combines the parameters of two parent policies to produce an offspring. EPO employs weighted averaging of parameters (i.e., the weights of the neural networks) with contributions from each parent determined by their relative fitness, $f_1$ and $f_2$:

\[
c = \alpha p_1 + (1 - \alpha) p_2,
\]
where
\[
\alpha = \frac{f_1}{f_1 + f_2 + \epsilon}.
\]

This fitness-weighted scheme biases the offspring's parameters towards the higher-performing parent, promoting exploitation of superior policies. The small constant \(\epsilon\) prevents division by zero and ensures numerical stability.

Although fitness-based crossover leverages the strengths of high-performing parents, it may reduce exploration by overly favoring dominant policies. Mutation, as described below, is used to counterbalance this effect. Other crossover techniques, such as random parameter masking, were tested but did not outperform the proposed weighted averaging approach. More sophisticated methods for performing crossover remain an area for further exploration in the future, but the current crossover method employed in EPO is sufficient to demonstrate the efficacy of combining policy gradients and neuroevolution, while requiring no hyperparameter tuning to work well in other domains.

\subsubsection{Mutation}
Mutation enhances exploration by introducing diversity into the offspring's parameters post-crossover. Gaussian noise is added adaptively to the network weights, with the magnitude of noise inversely proportional to the fitness difference between the parents:
\[
\text{scaling\_factor} = \max(0.01, \min(0.1, \frac{| f_1 - f_2|}{\max(f_1 + f_2, \epsilon)})).
\]
The noise is then sampled as:
\[
\text{mutation noise} \sim \mathcal{N}(0, \sigma^2),
\]
where \(\sigma^2\) is proportional to the scaling\_factor. This adaptive mechanism encourages significant exploration when parents have similar fitness levels (indicating less certainty in the optimal direction) while preserving exploitation when a clear fitness leader exists.

Mutation probability is a hyperparameter that controls the likelihood of introducing random noise into offspring parameters after crossover. A higher mutation probability promotes exploration by increasing diversity in the population, while a lower probability emphasizes exploitation by preserving the inherited traits of high-performing parents.

% MOVE AFTER EVOLUTIONARY OPERATIONS 
\subsection{Local Optimization with PPO}
PPO plays a dual role in EPO, serving both to pre-train the initial policy and to fine-tune selected offspring during the evolutionary process. Throughout, the Stable Baselines implementation of PPO is used with standard hyperparameters to ensure consistency and reproducibility \cite{RaffinStableBaselines3}.

\paragraph{Pre-Training the Initial Policy}
The initial population is derived from a policy pre-trained using PPO for a limited number of timesteps. The resulting pre-trained policy is then cloned to initialize the population, ensuring all agents begin evolution from a strong baseline. This approach avoids random initialization, which can hinder performance in high-dimensional policy spaces.

\paragraph{Fine-Tuning During Evolution}
During evolution, PPO is applied to refine offspring generated through crossover. Only offspring not subjected to mutation are fine-tuned, as these are more likely to inherit desirable traits from elite parents. Fine-tuning is computationally efficient, involving a limited number of gradient steps (e.g., 500 timesteps) per offspring. 

\begin{table*}[thb]
  \caption{Comparison of Benchmark Tasks}
  \label{tab:task_comparison}
  \begin{tabular}{lllll}
    \toprule
    Environment & Reward Density & Task Complexity & State Space & Key Challenges \\
    \midrule
    Pong & Dense & Low & Small & Paddle control, trajectory prediction \\
    Breakout & Medium & Medium & Moderate & Precise paddle control, targeting specific bricks \\
    \bottomrule
  \end{tabular}
\end{table*}
\subsection{Hyperparameter Tuning}

Hyperparameter optimization is critical for designing learning algorithms. Hyperparameters, such as learning rates and discount factors, shape an agent’s ability to explore, learn, and generalize. Poor tuning can slow learning or cause divergence, while well-optimized ones improve stability and sample efficiency. Hyperparameters, unlike model parameters, are not learned during training but need to be set before training begins. 

Searching for the optimal combination of hyperparameters is often challenging due to the complexity of the search space. Traditional search approaches are straightforward but inefficient, especially for high-dimensional spaces or computationally expensive evaluations. 

A simple yet effective way to optimize hyperparameters is Bayesian optimization. Bayesian optimization is a probabilistic model-based optimization technique, which offers a more efficient alternative to traditional optimization methods. The key idea is to model the objective function (e.g., algorithm performance) with a surrogate function (e.g., Gaussian Processes) and iteratively select hyperparameters based on this model to balance exploration (searching new regions) and exploitation (refining known good regions). 

EPO hyperparameters were optimized with Bayesian optimization using Optuna \cite{AkibaOptuna2019}. Optuna is an open-source hyperparameter optimization framework that implements advanced techniques such as Bayesian optimization to automate and accelerate the search for optimal hyperparameters. Optuna employs a tree-structured Parzen estimator (TPE) as its surrogate model, which balances exploration of new configurations and exploitation of known good ones. TPE is especially efficient for high-dimensional and complex hyperparameter spaces \cite{BergstraTPE2011}. 

Optimal hyperparameters were searched over 75 trials of 240 seconds of training on Breakout, with results averaged over five runs. Three hyperparameters were optimized: mutation probability, elite count, and population count.

Table \ref{tab:hyperparameters} provides an overview of each hyperparameter, including its description, the range of values explored during the search, and the optimal value identified. A summary of hyperparameter configurations with corresponding rewards can be found in the Appendix~\ref{appendix:hyperparam_optim}.

The optimal configuration identified was:
\begin{itemize}
    \item \textbf{Mutation Probability}: A mutation probability of $0.3$ emerged as optimal, striking a balance between exploration and stability. Higher mutation rates, such as $0.5$, introduced excessive randomness, leading to diminished performance. Conversely, lower mutation rates, such as $0.1$, restricted diversity, causing the evolutionary process to stagnate.
    
    \item \textbf{Population Count}: While larger populations, such as twelve, showed marginal improvements in performance due to increased exploration, they imposed a significant computational cost without proportional gains in reward. A population size of eight provided a good trade-off, maintaining diversity without compromising sample efficiency.
    
    \item \textbf{Elite Count}: Retaining three top-performing agents per generation proved effective in stabilizing the evolutionary process. A higher elite count of five led to reduced exploration, as the population became dominated by a small number of agents, whereas an elite count of two resulted in insufficient retention of high-quality individuals.
\end{itemize}

In 20 trials for Pong, with 240 seconds of training per trial averaged over 10 runs, EPO with the identified hyperparameters consistently performed well. This indicates that EPO requires minimal hyperparameter tuning across tasks, as the configuration generalizes effectively without task-specific adjustments.

\section{Experiments}
\label{sec:experiments}

 The Atari games domain is a common benchmark for evaluating the performance of RL algorithms on partially observed tasks with high-dimensional visual input \cite{MnihAtariDeepRL}. EPO's performance was evaluated on two games from the Atari benchmarks: Breakout and Pong. 
\begin{comment}
The goal was to determine whether EPO offers consistent performance improvements over PPO across different problem domains. These domains were selected to evaluate EPO's capacity to perform consistently across a range of tasks, from structured, low-dimensional problems to complex, high-dimensional environments.
\end{comment}

\subsection{Experimental Setup}

To ensure a fair comparison, all methods were trained and evaluated using identical computational resources. We used a single NVIDIA Tesla T4 GPU with 15 GB of video RAM, 51 GB of system RAM and 240 GB of memory. Training durations were capped at 10,000 seconds of wall-clock time for Pong and 7,200 seconds for Breakout per experiment for all methods to ensure comparability under resource constraints. 

Each experiment was repeated over 10 runs with different random seeds to account for stochasticity, and results were averaged across runs. PPO serves as the primary baseline, utilizing the Stable Baselines implementation with the default hyperparameter configuration.

\subsection{Atari Game Playing}

EPO was trained to learn policies on Pong and Breakout, which represent a range of challenges: Pong is a dense-reward, low-complexity task, while Breakout features sparse rewards and higher task complexity. In addition, the computational overhead of pre-processing these visual inputs makes Atari games a suitable benchmark for evaluating sample complexity. Table~\ref{tab:task_comparison} provides a summary of these tasks in terms of reward density, task complexity, state space, and challenges. 

\subsubsection*{Domain Setup}

Atari tasks were simulated using the Arcade Learning Environment \cite[ALE; ][]{BellemareALE}. Raw Atari frames, originally 210×160 pixel RGB images, were preprocessed to reduce dimensionality and capture temporal dynamics. Following standard practice \cite{Mnih2015HumanlevelCT}, frames were resized to 84×84, and the last four frames were stacked to provide temporal context. RGB input was used instead of converting it to grayscale to preserve richer visual features. Frame skipping was employed to enhance computational efficiency, retaining every fourth frame. To promote generalization and prevent overfitting to deterministic initial states, a random number of no-op actions are applied at the start of each episode, introducing variability into the environment's starting conditions.

The preprocessed frames were passed through a convolutional neural network (CNN) \cite{Lecun1989} to encode the input into a compact feature representation. The network consisted of three convolutional layers and one fully connected layer. The first convolutional layer used 32 filters of size 8×8 with a stride of four, the second 64 filters of size 4×4 with a stride of two, and the third 64 filters of size 3×3 with a stride of one. Each convolutional layer was followed by a ReLU activation. The final fully connected layer had 512 units and output both the policy and value function. The policy network computes action probabilities via a softmax activation, while the value function estimates the expected cumulative reward.

EPO's performance was evaluated on two metrics: reward progression over training time and reward distribution across episodes post-training. 

Reward over training time provides insights into learning stability and convergence behavior. This metric is particularly useful for understanding how quickly EPO adapts to the task compared to PPO. Reward distribution over episodes, measured after training concludes, evaluates the performance of the final policy across multiple episodes. Together, these metrics provide a comprehensive evaluation of both the learning process and the resulting policy.

An important further metric is the total sample count, which was used to compare the sample efficiency of each method.  Sample efficiency determines whether a policy optimization method is viable, particularly when environment interactions are costly. This metric is straightforward to measure for the end-to-end training process of PPO but requires additional tracking mechanisms for EPO. For EPO, the total sample count included the number of samples used for PPO pre-training, PPO fine-tuning, and fitness evaluations during the evolutionary process. 

\subsubsection*{Results for Pong}

\begin{figure}
    \centering
    \includegraphics[width=1\linewidth]{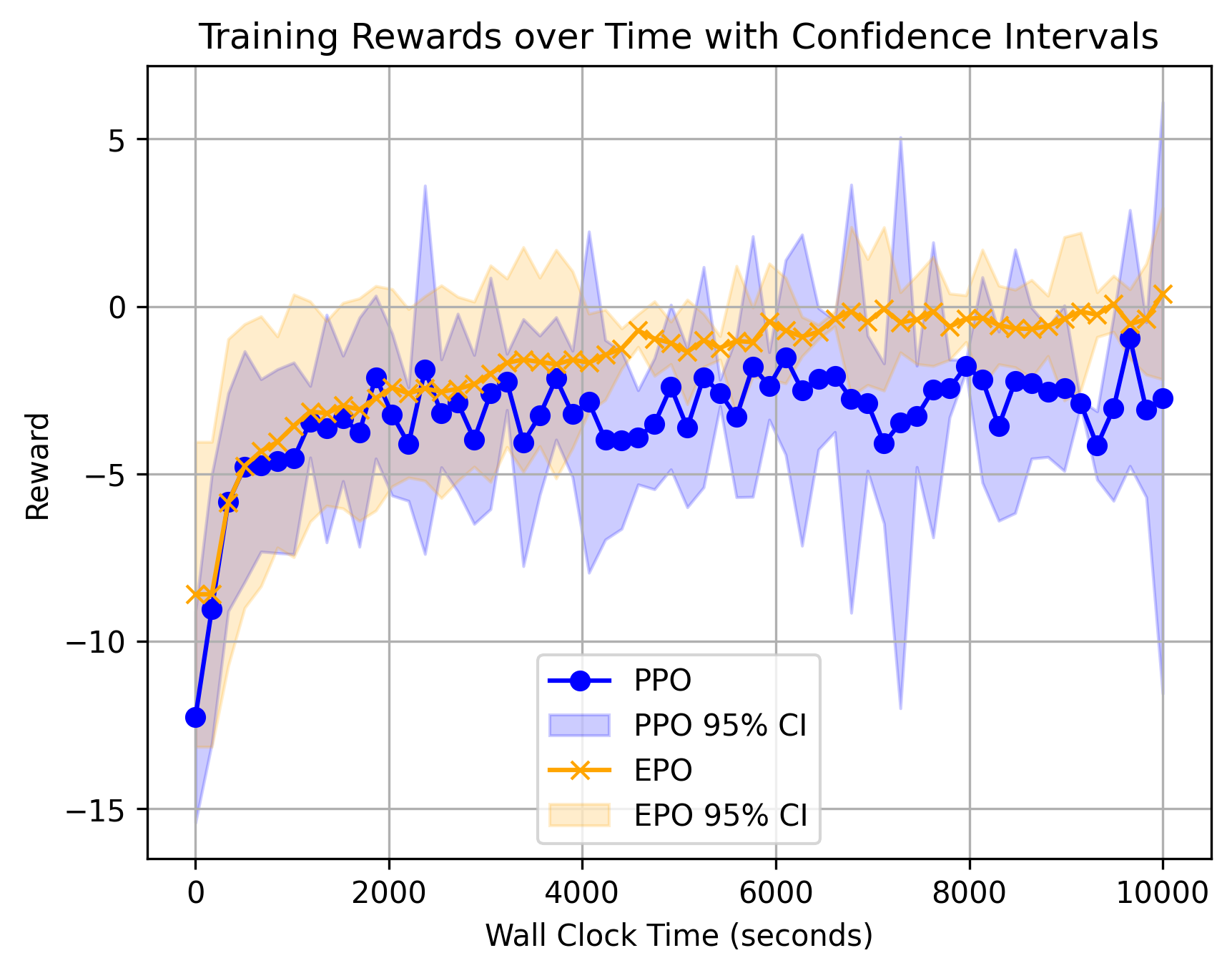}
    \caption{Training rewards over \textbf{10,000 seconds} of wall clock time for PPO and EPO on Pong with shaded areas indicating 95\% confidence intervals. EPO demonstrates steady and consistent improvement throughout the training process, with decreasing variance over time, indicating stable convergence. In contrast, PPO shows rapid early learning but stagnates later on, with larger variability in performance.}
    \label{fig:pong-training}
\end{figure}

Figure~\ref{fig:pong-training} shows the average rewards during training for EPO and PPO on Pong over 10,000 seconds. Overall, EPO was competitive with PPO and became better over time. Its confidence intervals became smaller and rewards steadily increased, while PPO peaked during the initial stages of training but then became stagnant.

Table~\ref{tab:pong_performance} compares sample complexity and average reward over twenty complete training episodes for PPO and EPO agents. EPO's mean reward is higher (\(-6.25 \pm 3.42\)) compared to PPO's (\(-8.03 \pm 5.17\)), reflecting a superior ability to learn effective policies. Additionally, the best reward achieved by EPO (\(11.0\)) far exceeds that of PPO (\(1.0\)), showcasing EPO's greater capacity for exploring high-reward policy regions. While EPO required more samples (\(2.36 \times 10^6\) vs. \(1.58 \times 10^6\)), this additional training translated into better policies and more consistent performance. The results highlight a tradeoff: EPO sacrifices some sample efficiency but gains a substantial improvement in policy quality.  

% Pong Table
\begin{table}
\caption{Performance Metrics Over 20 Episodes on Pong After 10,000 Seconds of Training}
\label{tab:pong_performance}
\centering
\small
\begin{tabular}{lcc}
\toprule
\textbf{Metric} & \textbf{PPO} & \textbf{EPO} \\
\midrule
Sample Count & $(1.58 \pm 0.01) \times 10^6$ & $(2.36 \pm 0.02) \times 10^6$ \\
Mean Reward  & $-8.03 \pm 5.17$ & $-6.25 \pm 3.42$ \\
Best Reward  & 1.0  & 11.0  \\
\bottomrule
\end{tabular}
\end{table}

\subsubsection*{Results for Breakout}

\begin{figure}
    \centering
    \includegraphics[width=1\linewidth]{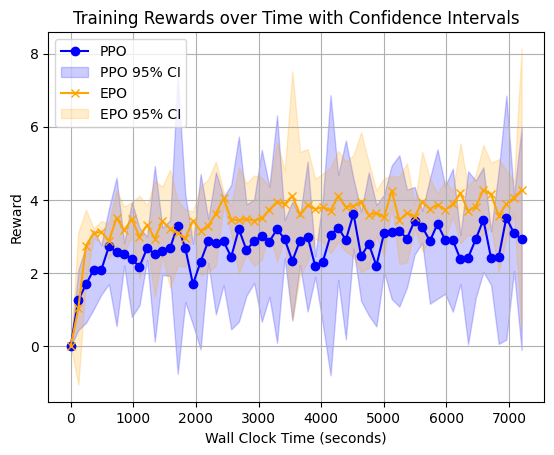}
    \caption{Training rewards over 7,200 seconds of wall clock time for PPO and EPO on Breakoutwith shaded areas indicating 95\% confidence intervals. EPO demonstrates a consistent upward trajectory in rewards with reduced variance over time. In contrast, PPO exhibits rapid early learning, but higher variance and stagnation in performance.}
    \label{fig:breakout-training}
\end{figure}

Figure~\ref{fig:breakout-training} shows the average rewards during training for EPO and PPO on Breakout over 7,200 seconds. EPO demonstrates consistently higher rewards throughout training, converging more quickly to stronger policies by exploring high-reward regions of the action space during evolution. While PPO shows more variability and slower progress, EPO maintains stability and achieves superior performance, as reflected in its higher reward curve and narrower confidence intervals at later stages.

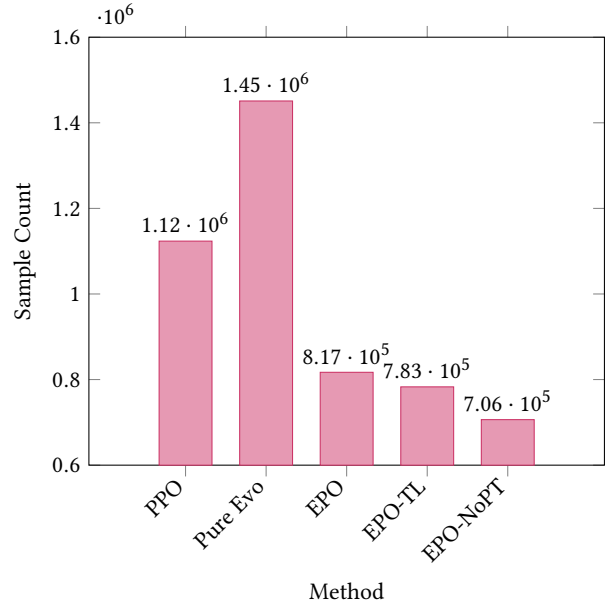
\begin{figure}
\centering
\caption{Sample complexity on Breakout across methods after 7,200 seconds of training. EPO and variants are much more sample efficient compared to PPO and pure evolution.}
\label{fig:breakout_sample_complexity}
\begin{tikzpicture}
    \begin{axis}[
        ybar,
        symbolic x coords={PPO, Pure Evo, EPO, EPO-TL, EPO-NoPT},
        xtick=data,
        ymin=600000, ymax=1600000,
        ylabel={Sample Count},
        xlabel={Method},
        bar width=20pt,
        nodes near coords,
        enlarge x limits=0.3,
        x tick label style={rotate=45, anchor=east},
        legend style={at={(0.5,-0.15)},anchor=north,legend columns=-1}
    ]
         \addplot[fill=purple!40,draw=purple!80]
         coordinates {(PPO,1123513.5) (Pure Evo,1450982.4) (EPO,816974.3) (EPO-TL,782962.0) (EPO-NoPT,706372.3)};
    \end{axis}
\end{tikzpicture}
\end{figure}

Table~\ref{tab:breakout_performance} compares sample complexity and average reward over twenty complete episodes for trained PPO and EPO agents, and Figure~\ref{fig:breakout_sample_complexity} visualizes the sample complexity for each method. EPO's rewards were comparable or slightly higher (\(3.15 \pm 1.45\)) than PPO's (\(3.02 \pm 1.43\)), but sample efficiency was significantly better. The reduction in sample count corresponds to a 26.8\% improvement compared to PPO, and a 57.3\% improvement compared to pure evolution. This improved sample efficiency on Breakout may stem from its greater demand for exploration compared to Pong, where EPO's evolutionary mechanisms are particularly effective. \\

% Breakout Table
\begin{table*}
\caption{Performance Metrics Over 20 Episodes on Breakout After 7,200 Seconds of Training}
\label{tab:breakout_performance}
\centering
\small
\begin{tabular}{lcccc}
\toprule
\textbf{Metric} & \textbf{PPO} & \textbf{EPO} & \textbf{EPO w/o Pre-training} & \textbf{Pure Evolution} \\
\midrule
Sample Count & $(1.12 \pm 0.002) \times 10^6$ & $(0.82 \pm 0.02) \times 10^6$ & $(0.71 \pm 0.02) \times 10^6$ & $(1.92 \pm 0.02) \times 10^6$ \\
Mean Reward  & $3.02 \pm 1.43$ & $3.15 \pm 1.45$ & $2.55 \pm 1.05$ & $0.40 \pm 0.47$ \\
Best Reward  & 7.0  & 7.0  & 6.0  & 2.0  \\
\bottomrule
\end{tabular}
\end{table*}

EPO outperformed existing methods by integrating exploration with exploitation in terms of performance and sample efficiency. Neuroevolution provides the key mechanism for effective exploration, making this combination particularly powerful.

\subsection{Ablation Study}
To analyze the contributions of pre-training and fine-tuning on EPO's performance on Breakout, ablation studies were conducted.

\begin{figure}
    \centering
    \includegraphics[width=1\linewidth]{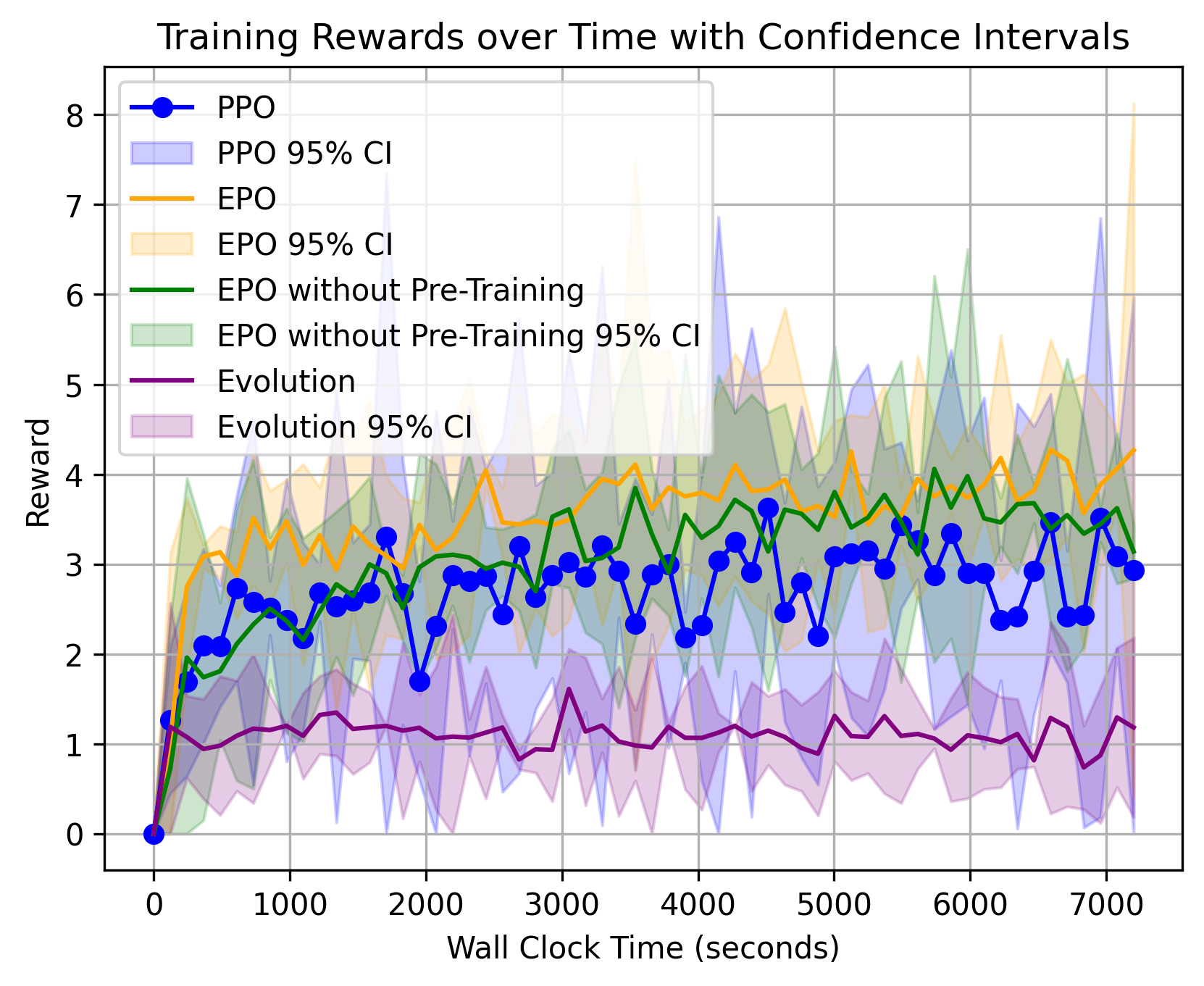}
    \caption{Training rewards over 7200 seconds of wall clock time for PPO, EPO, EPO without pre-training, and Evolution on Breakout with shaded regions representing 95\% confidence intervals. EPO without pre-training does not exceed EPO, but it outperforms PPO. Evolution without gradient-based optimization fails to achieve competitive performance and stagnates.}
    \label{fig:breakout-variants}
\end{figure}

Figure~\ref{fig:breakout-variants} shows the average rewards during training for PPO, EPO, and two EPO variants: EPO without pre-training, and pure evolution over 7,200 seconds of training. 

Pre-training in EPO accelerates the learning process, but does not significantly impact the accumulation of overall training rewards. Evolution struggles to accumulate rewards, indicating its limited ability to refine policies effectively on its own. The narrow confidence interval suggests that its performance was reliable, but it was consistently suboptimal.

\begin{figure}
    \centering
    \includegraphics[width=1\linewidth]{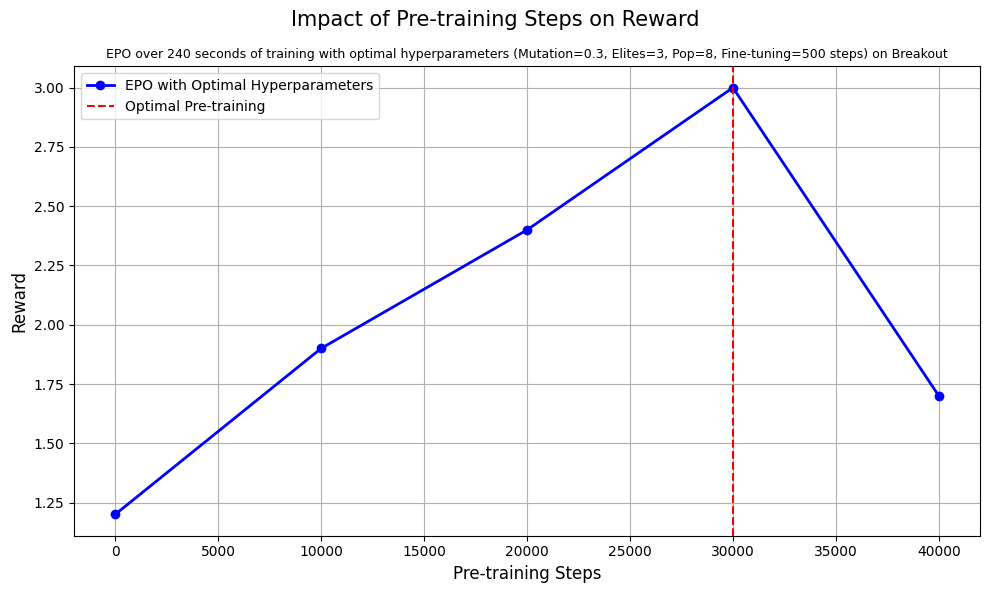}
    \caption{Training reward over 240 seconds across different pre-training steps for Breakout to illustrate the impact of the pre-training duration on training rewards.}
    \label{fig:pretraining-impact}
\end{figure}

Pre-training substantially influences initial performance, as illustrated in Figure~\ref{fig:pretraining-impact}. Without pre-training, the evolutionary process struggles to find high-performing regions, leading to suboptimal rewards and increased computational demands in initial stages. 

Even a small amount of pre-training improves evolution significantly but diminishing returns were observed beyond 30,000 steps, and additional pre-training adds computational cost without meaningful gains. At 40,000 steps, rewards begin to decline slightly, suggesting that excessive pre-training may overfit the initial population to specific regions of the parameter space, restricting the algorithm's exploration abilities. Thus, 30,000 steps was the optimal choice for pre-training to receive substantial gains while avoiding the computational inefficiency and potential overfitting observed under increased pre-training.

Fine-tuning was varied from 0 to 1000 steps in increments of 500 to understand the impact of local optimization on training rewards. While fine-tuning does not drastically improve final rewards, it reduces the number of generations needed to surpass PPO's baseline performance. This effect makes fine-tuning a valuable addition for scenarios where computational resources are constrained or faster convergence is desired. For applications without strict sample efficiency requirements, fine-tuning may be omitted without significant loss of performance.

\section{Transfer Learning}

Pre-training was shown to accelerate the evolutionary process in ablation experiments, demonstrating that initializing the population with a baseline agent that has undergone some training significantly improves performance. 

In previous experiments, these baselines were agents pre-trained for a small number of timesteps using PPO on the same task. An alternative approach involves leveraging a policy network trained on a different task to initialize the population for evolution, enabling the use of existing agents as a starting point and potentially reducing the need for task-specific training. Since Pong and Breakout are both games with similar actions involving paddles, one might wonder whether an agent trained on one game might learn the other game faster than an untrained agent. 

To evaluate the feasibility of transferring knowledge across tasks with this approach, an agent for Breakout was initialized using the weights of a policy network trained on Pong with EPO for 10,000 seconds. The performance of this transfer learning approach was compared to two baselines: PPO, and the EPO algorithm as described in Section~\ref{sec:epo}.

\begin{figure}
    \centering
    \includegraphics[width=1\linewidth]{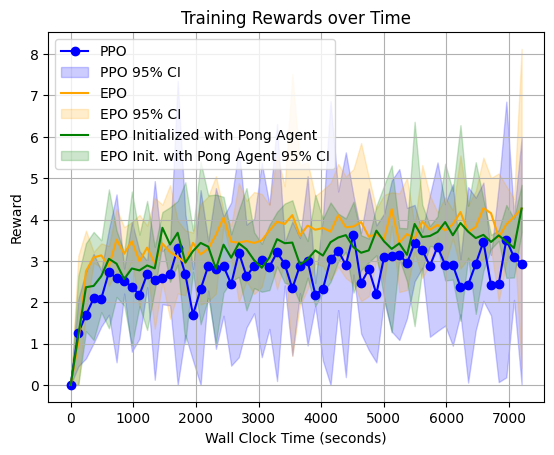}
    \caption{Training rewards over 7,200 seconds of wall clock time for PPO, EPO, and EPO-TL on Breakout with shaded regions representing 95\% confidence intervals. EPO-TL outperforms PPO, but fails to achieve competitive performance compared to EPO.}
    \label{fig:tl-training}
\end{figure}

Table~\ref{tab:epo_tl_performance} presents the sample complexity, and mean and best evaluation rewards for this transfer learning approach, denoted EPO-TL. Figure~\ref{fig:tl-training} shows the experimental results for EPO initialized with the weights of a Pong agent. 

Although training rewards are competitive compared to PPO, EPO-TL does not result in better training efficiency and final performance compared to EPO. In contrast, leveraging pre-trained weights from Pong led to a decrease in mean and best evaluation rewards compared to those seen in Section~\ref{sec:experiments}. These results suggest that although Pong and Breakout are both paddle games and one might expect transfer learning to work well, the policy spaces for the two games are sufficiently different that transfer learning was not useful in accelerating learning.

% EPO-TL Table
\begin{table}
\caption{Performance Metrics for EPO-TL on Breakout}
\label{tab:epo_tl_performance}
\centering
\begin{tabular}{lc}
\toprule
\textbf{Metric} & \textbf{EPO-TL} \\
\midrule
Sample Count     & 782,962.0 ± 26,791.7 \\
Mean Reward      & 2.07 ± 1.20                \\
Best Reward      & 6.0                \\
\bottomrule
\end{tabular}
\end{table}

\section{Discussion and Future Work}

The results of this study illustrate the effectiveness of EPO for policy optimization. Pre-training provides a strong starting point, while local optimization improves efficiency and enables the exploitation of top-performing agents. However, transfer learning between the two tasks used in the evaluation showed limited benefits.

In future work, EPO will be applied to more complex domains, such as the simulated robotics tasks in MuJoCo. Alternative population initialization strategies, and crossover and mutation schemes will be explored to better understand the role of population diversity in learning efficiency and policy performance.

EPO demonstrates the benefits of combining gradient-based and gradient-free approaches for deep RL tasks. EPO's applications in settings that require exploration without compromising performance, sample efficiency, and stability are promising.

\section{Conclusion}

This paper introduced Evolutionary Policy Optimization (EPO), a hybrid algorithm that combines the exploitation strength of policy gradients with the exploration capabilities of evolutionary methods. 

EPO was evaluated on two Atari benchmarks, Pong and Breakout. Compared to PPO and pure evolution, EPO demonstrated superior performance, achieving smoother and more stable training progress, and higher rewards over training and evaluation on both tasks. EPO also demonstrated increased sample efficiency on Breakout.

These results suggest that EPO can address reinforcement learning challenges in situations in which traditional methods struggle, particularly in environments requiring aggressive exploration and exploitation. 

\bibliographystyle{ACM-Reference-Format}
\bibliography{sample-base}

\appendix

\section{Hyperparameter Optimization}
\label{appendix:hyperparam_optim}

Table \ref{tab:hyperparameter-configs} shows the values of a subset of the hyperparameter configurations and the corresponding mean reward over 240 seconds of training on Breakout, averaged over 10 runs. These configurations were chosen to illustrate trends in performance and justify the selection of the optimal values. 

\begin{table}[h]
\centering
\caption{Hyperparameter Configuration Examples with with Corresponding Rewards (Over 240s of Training)}
\label{tab:hyperparameter-configs}
\begin{tabular}{cccc}
\toprule
\textbf{Mutation Prob.} & \textbf{Elite Count} & \textbf{Population Size} & \textbf{Reward} \\
\midrule
         \textbf{0.30} & \textbf{3} & \textbf{8} & \textbf{3.00} \\
         0.40 &                5 &                8 &       2.56 \\
         0.40 &                3 &               12 &       2.50 \\
         0.40 &                3 &               10 &       2.40 \\
         0.30 &                3 &               10 &       2.00 \\
         0.30 &                2 &               12 &       2.00 \\
         0.40 &                3 &                8 &       2.00 \\
         0.30 &                2 &                8 &       1.80 \\
         0.20 &                2 &                8 &       1.80 \\
         0.10 &                5 &               12 &       1.00 \\
\bottomrule
\end{tabular}
\end{table}

\end{document}